\documentclass[11pt,a4paper]{article}

\usepackage[final]{acl2024}
\usepackage{times}
\usepackage{latexsym}
\usepackage{graphicx}
\usepackage{booktabs}
\usepackage{amsmath,amssymb}
\usepackage{hyperref}
\usepackage{xcolor}
\usepackage{listings}
\usepackage{caption}
\usepackage{enumitem}
\usepackage{array}
\usepackage{tabularx}
\usepackage{xurl}

\hypersetup{
    colorlinks=true,
    linkcolor=blue,
    filecolor=magenta,
    urlcolor=blue,
    citecolor=blue,
    pdftitle={ArguLens: An Open-Source System for Automated Essay Scoring and Label-Aware Feedback Generation},
    pdfauthor={Weiran Wang; Hongxiang Shi; Huitao Tang; Wenjuan Qin},
}

\title{%
\includegraphics[height=0.34in]{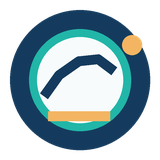}\\[-0.1em]
ArguLens: An Open-Source System for\\
Automated Essay Scoring and Label-Aware Feedback Generation}

\author{
    Weiran Wang\textsuperscript{*} \quad Hongxiang Shi \quad
    Huitao Tang \quad Wenjuan Qin\\
    Fudan University\\
    \texttt{wrwang25@m.fudan.edu.cn}\\
    \textsuperscript{*} Corresponding author
}

\date{\today}

\begin{document}
\maketitle

\begin{abstract}
Most automated essay scoring (AES) systems output a single holistic score
without interpretable evidence and rely on closed APIs that introduce data
privacy and cost barriers. We present \textbf{ArguLens}, an
open-source, locally deployable system that decomposes AES into three decoupled
components: a discourse-move classifier (Qwen2.5-7B-Instruct fine-tuned with
LoRA on PERSUADE 2.0), a grade-independent LightGBM scorer over 31
linguistic and discourse features, and a label-aware feedback generator
served through vLLM with a Qwen2.5-14B-Instruct backbone. A Gradio web UI
exposes pluggable inference backends and supports single-essay and batch
scoring with downloadable per-essay breakdowns. On an essay-disjoint
PERSUADE 2.0 test split, the logit-probe classifier achieves 82.6\%
accuracy and 0.727 macro-F1; under prompt-grouped 5-fold cross-validation
the scorer reaches a mean QWK of 0.813 under an oracle discourse-feature
protocol, and an ablation shows that adding gold discourse annotations yields
an increment of +0.055 QWK over the lexical+syntactic configuration (paired
$t$-test, $p=0.010$). This is a component-level diagnostic rather than an
end-to-end classifier-to-scorer result. The feedback generator ships with a
structured evaluation protocol; its human-rater study is left to future work.
The system is released under Apache 2.0 at
\url{https://github.com/wwrwbs/AI\_AWE}.
\end{abstract}

\section{Introduction}
\label{sec:intro}

Automated essay scoring (AES) has evolved from linear regression on surface
features \citep{page1966imminence} to deep neural architectures
\citep{taghipour2016neural,mayfield2020should} and, most recently, to
LLM-based holistic scoring \citep{beigman-klebanov-madnani-2020-automated,
tate2024holistic}. Despite this progress,
a fundamental tension persists between interpretability and performance.
Classic feature-based methods offer transparent scoring but cap ceiling
accuracy; neural models improve accuracy at the cost of opacity, and LLM
prompting requires closed APIs that restrict local deployment and data
privacy.

Three specific gaps remain for research and assistive deployment in
educational contexts. First, end-to-end models output a single score without
interpretable linguistic or rhetorical evidence, making it difficult for
instructors to understand how a grade was reached. Second, reliance on
commercial APIs introduces cost, latency, and data-privacy barriers that
limit adoption in resource-constrained settings. Third, most AES systems
stop at a holistic score; fine-grained, label-aware revision suggestions
that could help learners improve their writing are rare.

We address these gaps with \textbf{ArguLens}, a locally deployable,
Apache 2.0-licensed system whose design and evaluation are presented in this
paper. The system:
\begin{itemize}
    \item decomposes scoring into discourse-move classification, feature
    extraction, LightGBM scoring, and label-aware feedback;
    \item supports local deployment via vLLM with tensor parallelism and a
    4-bit HuggingFace fallback, while retaining an optional API backend;
    \item ships reproducible artifacts including the scorer model, scaler,
    LoRA adapter configuration in the repository, and weights as a versioned
    Release asset with checksums;
    \item exposes a config-driven stack with zero hard-coded paths,
    environment-variable precedence, and an offline test suite.
\end{itemize}

We evaluate the classifier and scorer on the PERSUADE 2.0 corpus
\citep{crossley2024persuade} of US middle-school argumentative essays graded
1 to 6. The discourse-move classifier is assessed through sentence-level
accuracy and macro-F1; the scorer is evaluated through quadratic weighted
kappa (QWK), accuracy, and macro-F1 under prompt-grouped cross-validation.
The feedback component is evaluated here at the level of its documented
generation workflow and repository contract; human-rater results are
explicitly outside the scope of this release.

\section{Related Work}
\label{sec:related}

\paragraph{AES Paradigms.}
Automated writing evaluation has a long history of balancing predictive
accuracy, interpretability, and instructional use
\citep{beigman-klebanov-madnani-2020-automated}. Classic feature engineering
\citep{kyle2015automatically,lu2010automatic} exposes linguistic evidence,
whereas neural AES models \citep{taghipour2016neural,mayfield2020should} can
learn richer representations at the cost of transparency. Recent work also
shows that general-purpose AI can provide useful holistic scoring, while
raising questions about rubric alignment and validity
\citep{tate2024holistic}. Our grade-independent LightGBM scorer retains an explicit
feature space while the LoRA classifier adds rhetorical structure to it.

\paragraph{Argument Structure and Feedback.}
Argument mining in persuasive and student essays has modeled both component
structure and essay quality \citep{persing2010modeling,stab2017parsing}.
Recent work further reports that combining argument-segment and cohesion
features improves automatic feedback-related scoring
\citep{ding-etal-2024-argumentation}. Recent benchmarking makes the
granularity issue explicit: EssayJudge evaluates AES at lexical, sentence,
and discourse levels and still finds substantial gaps at the discourse level
\citep{su-etal-2025-essayjudge}. We adopt the four PERSUADE 2.0 labels
(claim, data, counterclaim, rebuttal) and fine-tune a Qwen2.5-7B LoRA,
avoiding full-parameter updates while preserving the base model's language
understanding.

\paragraph{Feedback Evaluation and Fairness.}
Feedback quality is multidimensional rather than reducible to a single
automatic score. LLM-Rubric demonstrates the value of explicit dimensions and
calibration against human judgments when evaluating generated text
\citep{hashemi-etal-2024-llm}. More recent writing-feedback evaluation finds
that models may produce specific comments while missing the most important
problem or misjudging whether criticism is appropriate
\citep{rashkin-etal-2025-help}. Accordingly, this release defines relevance,
actionability, and tone as evaluation dimensions but does not report human
ratings. Fairness is also not implied by aggregate AES accuracy: subgroup
performance can vary with demographic and learner characteristics
\citep{schaller-etal-2024-fairness}. We therefore treat subgroup analysis as a
required future study rather than making a fairness claim from the aggregate
metrics reported here.

\paragraph{Open Educational AI.}
Our release follows model-reporting practice by documenting the model,
training context, limitations, and asset locations
\citep{mitchell2019modelcards}. Small artifacts are kept in the repository,
while the larger adapter is distributed as a versioned Release asset with a
checksum. This is a release design choice, not an experimental finding.

\section{System Architecture}
\label{sec:architecture}

\begin{figure*}[t]
\centering
\includegraphics[width=\textwidth]{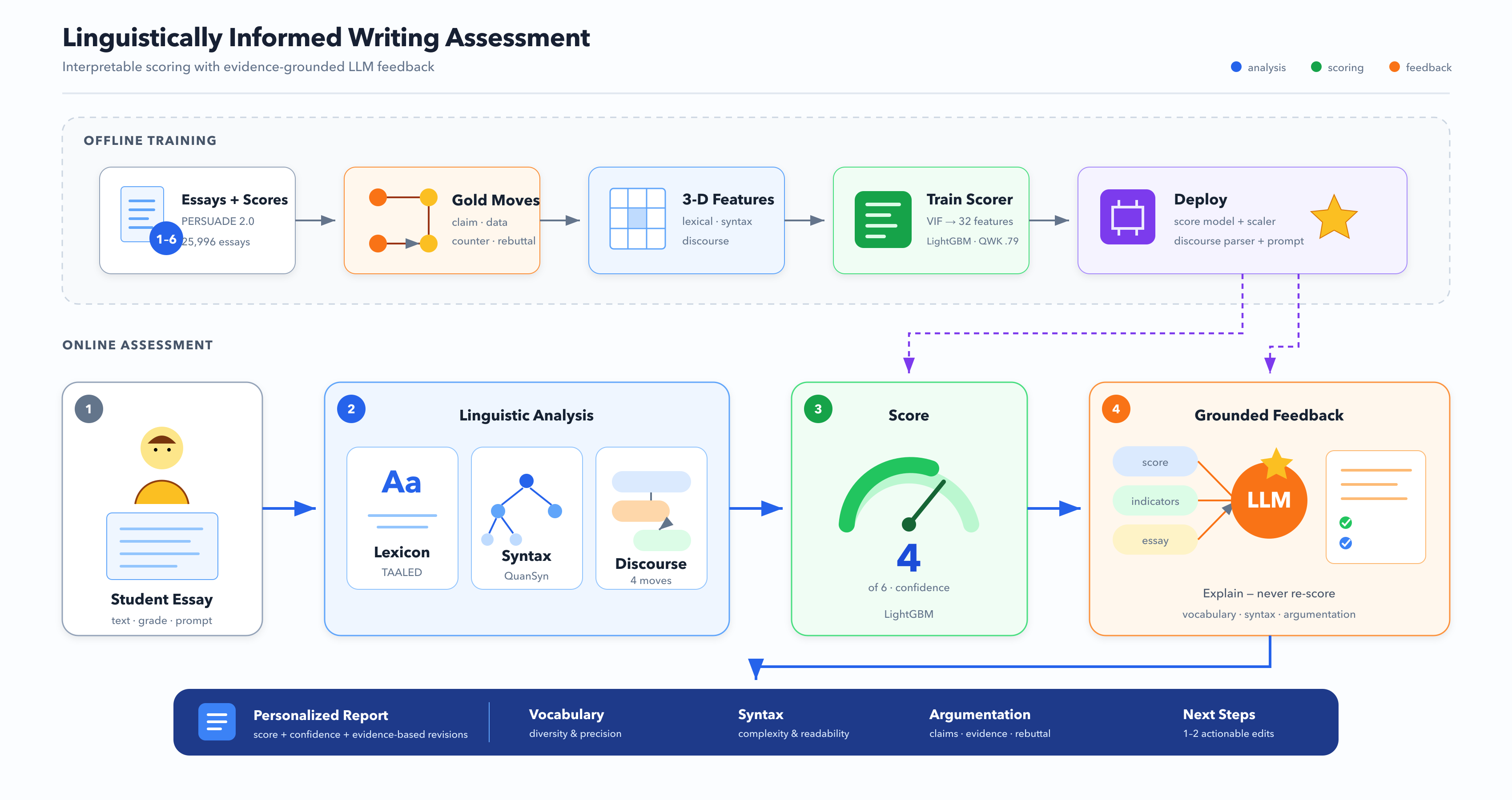}
\caption{Project workflow for linguistically informed writing assessment.
The upper lane summarizes offline training and deployment; the lower lane
shows online assessment, scoring, grounded feedback, and the personalized
report.}
\label{fig:architecture}
\end{figure*}

Figure~\ref{fig:architecture} illustrates the end-to-end pipeline. The
Gradio frontend (\texttt{essay\_score/app\_gradio.py}) orchestrates three
independent services whose responsibilities are summarised in
Table~\ref{tab:modules}. The move classifier runs through pluggable HF or
vLLM backends and emits sentence labels; the feature extractor combines
TAALED lexical indices, QuanSyn dependency metrics, and discourse counts
into a 31-feature vector; the scorer outputs a score, confidence, and
flags; and the feedback generator produces structured feedback through
vLLM or an OpenAI-compatible API.

\begin{table}[h]
\centering
\caption{Core modules and their responsibilities.}
\label{tab:modules}
\small
\begin{tabularx}{\columnwidth}{@{}p{0.24\columnwidth}p{0.25\columnwidth}X@{}}
\toprule
\textbf{Module} & \textbf{Path} & \textbf{Responsibility} \\
\midrule
Discourse-Move Classifier & \path{essay_score/infer} &
Sentence-level 4-way classification (LoRA on Qwen2.5-7B) \\
Scorer & \path{essay_score/scoring_pipeline} &
31-feature LightGBM multiclass (scores 1--6) \\
Feedback Generator & \path{essay_score/feedback} &
Label-aware English feedback via Qwen2.5-14B (vLLM) \\
Frontend & \path{essay_score/app_gradio.py} &
Gradio UI, batch processing, and ZIP export \\
Training Code & \path{qwen_move_classifier} &
LoRA SFT with LLaMA-Factory, data prep \\
\bottomrule
\end{tabularx}
\end{table}

\subsection{Configuration and Reproducibility}
\label{ssec:config}

All runtime settings follow a strict precedence: environment variables
override \texttt{config.yaml}, which in turn overrides default HuggingFace
model IDs. The configuration loader (\texttt{essay\_score/config.py})
resolves relative paths against the repository root. A
\texttt{model\_checksums.sha256} manifest enables \texttt{sha256sum -c}
verification of the scorer, scaler, adapter config, tokenizer, and LoRA
weights, ensuring that downstream users can verify asset integrity.

\section{Discourse-Move Classifier}
\label{sec:classifier}

\subsection{Task Formulation}

Given an essay split into $N$ sentences $\{s_1,\dots,s_N\}$, the classifier
predicts a label $y_i \in \{\texttt{claim},\texttt{data},
\texttt{counterclaim},\texttt{rebuttal}\}$ for each sentence. It operates at
sentence level with local context derived from the preceding sentence and the
essay opening.

\subsection{Base Model and LoRA}

The classifier uses the Apache-2.0-licensed
\texttt{Qwen/Qwen2.5-7B-Instruct} model \citep{qwen25} as its base. A LoRA
adapter \citep{hu2021lora}
with rank $r=32$, $\alpha=64$, and dropout $0.05$ is applied to all linear
projection modules (\texttt{q\_proj}, \texttt{k\_proj}, \texttt{v\_proj},
\texttt{o\_proj}, \texttt{gate\_proj}, \texttt{up\_proj}, \texttt{down\_proj}).
Training was conducted with LLaMA-Factory \citep{llamafactory} on PERSUADE 2.0
using 4-GPU DDP (\texttt{torchrun}) and BF16 precision, with a peak learning
rate of $1\times10^{-4}$ under a cosine schedule (warmup ratio $0.05$),
weight decay $0.01$, a cutoff length of 2{,}048 tokens, two epochs, and an
effective batch size of 32 (4 GPUs $\times$ batch 1 $\times$ gradient
accumulation 8). The repository ships the adapter configuration files; the
weight file (\texttt{adapter\_model.safetensors}, 309 MB) is published as a
GitHub Release asset tagged \texttt{v0.1.0}.

\subsection{Training Data and Model Selection}
\label{ssec:split}

PERSUADE 2.0 does not prescribe an official split, so we created an
essay-level split with seed 42: 22{,}605 essays for training, 2{,}825 for
validation, and 2{,}826 for testing, with zero essay overlap between all
pairs of splits (verified programmatically). Because the natural label
distribution is heavily skewed (\texttt{data} and \texttt{claim} dominate),
the training set is class-balanced by resampling to 25{,}000 sentences per
label (100{,}000 examples in total). A small balanced validation set
(2{,}000 sentences, 500 per label) monitors training, and the checkpoint
with the lowest validation loss is selected for evaluation; all reported
test results use the natural-distribution test set of 12{,}000 sentences
from 2{,}730 essays.

\subsection{Logit-Based Classification}

Rather than generating free text, the classifier uses a discriminative logit
probe for deterministic inference. For each target sentence, a chat-template
prompt is built with a system definition of the four move types encoded as
single-token codes (A, B, C, D). A forward pass through the base model with
the attached LoRA produces logits at the last non-padding position. The four logits corresponding to the code tokens are extracted and passed
through softmax; the $\arg\max$ is the predicted label and the maximum
softmax probability is the per-sentence confidence score. The HF backend
chunks prompts using a configurable sentence batch size, while the vLLM
backend sends a bounded request to a persistent service with one-token
decoding and the four allowed label tokens.

\subsection{Inference Backends}

Three backends are available. The \texttt{hf} backend loads the base model
and adapter in-process using \texttt{transformers} and \texttt{peft}. The
\texttt{vllm} backend runs a persistent FastAPI service that loads the base
model once and attaches the LoRA via \texttt{LoRARequest}
\citep{kwon2023vllm}. Its
\texttt{/health} endpoint reports the adapter rank, tensor parallelism size,
and GPU count. The frontend falls back to \texttt{hf} if the vLLM service is
not ready. The \texttt{api} backend is reserved for the feedback generator.

\section{Grade-Independent LightGBM Scorer}
\label{sec:scorer}

\subsection{Feature Set}

All features are computed without learner grade information to prevent direct
target leakage. This design does not establish fairness, because linguistic
features can still correlate with demographic or curricular variables.
Extraction uses a single shared spaCy dependency parse for both lexical and
syntactic metrics.

\paragraph{Lexical Features (TAALED \citep{kyle2015automatically}).}
Sixteen lexical features comprise twelve lexical-diversity variants across
AW, content-word, and function-word classes, together with
\texttt{word\_count}, \texttt{awl},
\texttt{basic\_nfunction\_tokens}, and \texttt{lexical\_density\_tokens}.

\paragraph{Syntactic Features (QuanSyn \citep{quansyn}).}
Eight dependency-based features are retained from QuanSyn:
\texttt{mdd}, \texttt{ndd}, \texttt{mhd}, \texttt{mtdl}, \texttt{vk},
\texttt{mtw}, \texttt{hi}, and \texttt{mrd}.

\paragraph{Discourse Features (deployment and evaluation protocols).}
The deployment pipeline derives seven features from the Qwen2.5-7B LoRA
move-classifier output: \texttt{n\_sent}, \texttt{n\_claim}, \texttt{n\_data},
\texttt{pct\_data}, \texttt{pct\_counterclaim}, \texttt{pct\_rebuttal}, and
\texttt{has\_counter\_rebut} (binary, true only when both a counterclaim and
a rebuttal are present). The scorer feature table used for the reported
cross-validation instead contains the corresponding PERSUADE gold discourse
annotations. We therefore call the reported scorer results an oracle
component evaluation and do not present them as an end-to-end classifier
benchmark.

\paragraph{Handling Missing Dependencies.}
If the \texttt{quansyn} package is unavailable, the scorer raises a
\texttt{FeatureExtractionError} by default. A configuration flag
(\texttt{allow\_missing\_dependency\_features}) fills missing values with
zeros for exploratory runs only.

\subsection{Model Training}

The scorer uses LightGBM \citep{ke2017lightgbm} with the \texttt{multiclass}
objective over six ordinal classes. Hyperparameters include
\texttt{num\_leaves=63}, \texttt{learning\_rate=0.03},
\texttt{n\_estimators=1000}, \texttt{min\_child\_samples=15},
\texttt{subsample=0.8}, \texttt{colsample\_bytree=0.8}, and
\texttt{reg\_lambda=0.1}. Training uses a seed of 42; each cross-validation
fold increments the model seed by its fold index. Training data are the
PERSUADE 2.0 grade-independent feature table, with prompt groups held out during
cross-validation.
The trained model (\texttt{lightgbm\_multiclass.txt}, 33 MB) and the feature
scaler (\texttt{scaler.json}) are committed to the repository.

At inference, raw feature values are z-score normalized and passed to the
LightGBM model, which outputs a six-class probability vector. The expected
value is rounded to the nearest integer in 1--6 to produce the final score.
The same expected-value decoding rule is used by cross-validation, ablation,
and production inference. The reported cross-validation uses the
\texttt{gold\_annotations} discourse-feature source; production replaces
those fields with predicted move labels from the classifier.
The system flags essays with \texttt{low\_confidence} when the maximum class
probability falls below 0.5 and marks them as \texttt{borderline} when the
runner-up probability exceeds 0.30 and is adjacent to the predicted class;
both thresholds are fixed heuristics rather than tuned values.

\section{Label-Aware Feedback Generation}
\label{sec:feedback}

\subsection{Prompt Design}

Feedback is generated on every analysis. The prompt template
(\path{feedback/prompt_essay_v2.json}) receives the essay text, the essay
prompt name, the predicted score, confidence, and flags, together with a
curated subset of 14 of the 31 features---four lexical (e.g., AWL, MTLD),
five syntactic (MDD, MHD, MRD, MTDL, head-initial ratio), three discourse
(claim and evidence counts, counterclaim/rebuttal indicator), and two
length features. Each feature is accompanied by a band label (Low, Medium,
or High) derived from approximate corpus-level PERSUADE 2.0 tertiles; for
example, an AWL below 4.22 is labelled Low, between 4.22 and 4.54 Medium,
and above 4.54 High, while MDD, MHD, MRD, and MTDL use lower-is-better
scoring. These bands are used only for prompting the feedback model and
never enter the scorer, so they cannot leak score information. The system
prompt enforces constructive, English-only output, forbids inventing
metrics, requires quoting real essay text, and instructs the model to
recommend human review whenever the low-confidence or borderline flag is
set. The output contract fixes exactly five sections: overall assessment,
lexical dimension, syntactic dimension, discourse/argumentation dimension,
and the top one or two actionable improvements.

\subsection{Backends}

Three backends are available. The \texttt{vllm} backend (default) runs a
persistent OpenAI-compatible server (port 8000, TP=2 for the 14B model). The
\texttt{hf} backend uses 4-bit \texttt{BitsAndBytesConfig} with SDPA on a
single GPU. The \texttt{api} backend accepts any OpenAI-compatible endpoint;
the API key is read from the UI or environment at runtime and is never stored.

\section{Frontend, Batch Processing, and Export}
\label{sec:frontend}

The Gradio 6.x interface (\texttt{essay\_score/app\_gradio.py}) provides
single-essay analysis, batch upload with summary table, and ZIP export. Each
per-essay folder in the archive contains \texttt{essay.txt},
\texttt{score.json}, \texttt{linguistic\_features.json},
\texttt{discourse\_features.json}, \texttt{feedback.md},
\texttt{result.json}, and \texttt{sentence\_labels.csv}. The root contains
\texttt{summary.csv} and \texttt{manifest.json}.

\section{Experiments and Evaluation}
\label{sec:experiments}

We evaluate the classifier and scorer using distinct protocols because the
available artifacts are at different granularities. The classifier uses the
essay-disjoint sentence-level test split described in
Section~\ref{ssec:split}; the scorer uses prompt-grouped cross-validation on
the grade-independent feature table. We do not combine these protocols into
a single end-to-end score. Because the grouping variable is the essay
prompt, each scorer fold withholds complete prompts rather than randomly
splitting essays; this is a cross-prompt estimate, not an iid random-split
estimate. The released feature table does not include the raw essay text
needed to regenerate predicted discourse features for all scorer essays, so
the reported scorer numbers use gold discourse annotations and are explicitly
an oracle component evaluation.

\subsection{Experimental Setup}

\paragraph{Dataset.}
PERSUADE 2.0 \citep{crossley2024persuade} contains more than 25,000
argumentative essays written by US students in grades 6--12 across 15 prompts;
each essay is scored holistically on a 1--6 rubric. The corpus is available
under a non-commercial Creative Commons license from
\href{https://github.com/educational-technology-collective/persuade_2.0}{PERSUADE 2.0 repository}.

\paragraph{Metrics.}
For the discourse-move classifier, we report sentence-level accuracy,
macro-F1, and weighted-F1. For the LightGBM scorer, we report quadratic
weighted kappa (QWK) between predicted and human-assigned scores and
classification accuracy, both with standard deviations over the five
prompt-grouped folds. A protocol for evaluating the feedback generator on
relevance, actionability, and tone is defined in the repository;
human-annotator results are out of scope for this report. Inference latency
is reported as median and 95th percentile (p95) over 100 independent runs on
the reference hardware (1$\times$RTX 5000 Ada, 32 GB).

\paragraph{Ablation design.}
For the scorer, we ablate the 31-feature model against two reduced feature
sets: (1) the sixteen lexical features only (no discourse or syntactic
features); (2) lexical plus eight syntactic features (no discourse features).
The delta between the full model and configuration~(2) measures the marginal
value of the seven gold discourse features. It is an oracle diagnostic, not a
measurement of the downstream classifier's gain. No additional baseline was
run in this release; comparisons against ordinal
logistic regression, zero-shot or generative prompts, and fine-tuned encoder
classifiers remain future work rather than unreported results.

\subsection{Classifier Evaluation}
\label{ssec:classifier-results}

Table~\ref{tab:classifier-results} reports the discourse-move classifier
performance on the PERSUADE 2.0 test set, and Table~\ref{tab:confusion}
gives the corresponding confusion matrix. The \texttt{data} class achieves
the highest F1 (0.862), followed by \texttt{claim} (0.805). The two
classes with the largest error mass are \texttt{claim} and \texttt{data}:
confusion between them accounts for 1{,}528 of the 2{,}091 misclassified
sentences (73\%), reflecting the genuine difficulty of separating evidence
sentences from claims that embed evidence. The minority classes
\texttt{counterclaim} (F1 0.604) and \texttt{rebuttal} (F1 0.638) show low
precision (0.501 and 0.533) but comparatively high recall (0.761 and
0.794), a pattern consistent with the class-balanced resampling used
during training. Support counts are reported so that the macro-F1 is not
misread as performance on a balanced test set.

\begin{table}[t]
\centering
\caption{Discourse-move classifier results on the PERSUADE 2.0 test set
(12,000 sentences from 2,730 essays).}
\label{tab:classifier-results}
\scriptsize
\begin{tabularx}{\columnwidth}{@{}X>{\centering\arraybackslash}p{0.14\columnwidth}>{\centering\arraybackslash}p{0.13\columnwidth}>{\centering\arraybackslash}p{0.13\columnwidth}>{\centering\arraybackslash}p{0.13\columnwidth}@{}}
\toprule
\textbf{Label} & \textbf{Precision} & \textbf{Recall} & \textbf{F1} &
\textbf{Support} \\
\midrule
claim & 0.776 & 0.835 & 0.805 & 4,209 \\
data & 0.903 & 0.825 & 0.862 & 7,186 \\
counterclaim & 0.501 & 0.761 & 0.604 & 343 \\
rebuttal & 0.533 & 0.794 & 0.638 & 262 \\
\midrule
\multicolumn{5}{@{}l}{Overall: accuracy 82.58\%, macro-F1 0.727,
weighted-F1 0.830} \\
\bottomrule
\end{tabularx}
\end{table}

\begin{table}[t]
\centering
\caption{Confusion matrix on the PERSUADE 2.0 test set (rows: gold label;
columns: prediction).}
\label{tab:confusion}
\scriptsize
\begin{tabular}{@{}lcccc@{}}
\toprule
 & \textbf{claim} & \textbf{data} & \textbf{counter.} & \textbf{rebut.} \\
\midrule
claim & \textbf{3,515} & 564 & 81 & 49 \\
data & 964 & \textbf{5,925} & 171 & 126 \\
counterclaim & 25 & 50 & \textbf{261} & 7 \\
rebuttal & 23 & 23 & 8 & \textbf{208} \\
\bottomrule
\end{tabular}
\end{table}

\subsection{Scorer Evaluation}
\label{ssec:scorer-results}

Table~\ref{tab:scorer-results} reports the LightGBM scorer performance.
Fold-level QWK for the full model ranges from 0.754 to 0.872, indicating
heterogeneous difficulty across withheld prompts.

\begin{table}[t]
\centering
\caption{Scorer performance: QWK and accuracy (mean and standard deviation
over prompt-grouped 5-fold cross-validation on the 25{,}996-essay
grade-independent feature table).}
\label{tab:scorer-results}
\scriptsize
\begin{tabularx}{\columnwidth}{@{}X>{\centering\arraybackslash}p{0.22\columnwidth}>{\centering\arraybackslash}p{0.22\columnwidth}@{}}
\toprule
\textbf{Model} & \textbf{QWK $\uparrow$} & \textbf{Acc.\ (\%) $\uparrow$} \\
\midrule
\multicolumn{3}{@{}l}{\textit{Our approach}} \\
~~ LightGBM (all 31 features) &
  0.813 (0.048) & 62.73 (2.44) \\
\midrule
\multicolumn{3}{@{}l}{\textit{Ablations}} \\
~~ LightGBM (lexical only, 16 feats) &
  0.750 (0.071) & 57.57 (2.16) \\
~~ LightGBM (lexical + syntactic, 24 feats) &
  0.759 (0.069) & 58.69 (2.32) \\
\bottomrule
\end{tabularx}
\end{table}

\paragraph{Ablation: oracle discourse features.}
Adding gold discourse features yields +0.055 QWK over lexical+syntactic
features alone (0.813 vs.\ 0.759) and +0.064 over lexical-only features
(0.813 vs.\ 0.750). Because the five folds are prompt-aligned across
configurations, we can pair them: all five per-fold differences are
positive, with a mean paired difference of +0.057 (full vs.\
lexical+syntactic; paired $t(4)=4.64$, $p=0.010$) and +0.064 (full vs.\
lexical-only; paired $t(4)=5.04$, $p=0.007$). A Wilcoxon signed-rank test
cannot fall below $p=0.0625$ with five pairs, so we report the $t$-test
and note that these results should be read as a descriptive, component-level
diagnostic rather than conclusive evidence about predicted classifier
features in deployment.

\subsection{Feedback Generation and Evaluation Protocol}
\label{ssec:feedback-protocol}

The feedback generator produces label-aware, constructively framed revision
suggestions from the predicted discourse-move labels and feature bands (see
Section~\ref{sec:feedback}). A structured evaluation protocol of feedback
quality (relevance, actionability, and tone) is defined in the repository;
human-annotator results are left for future work and are not part of this
report.

\subsection{Latency Benchmarks}
\label{ssec:latency}

Table~\ref{tab:latency} reports component inference latency measured on a
single RTX 5000 Ada (32 GB VRAM) over 100 independent runs; it is not a
full-pipeline latency measurement.

\begin{table}[h]
\centering
\caption{Inference latency by component and backend. Hardware:
1$\times$RTX 5000 Ada (32 GB); 100 runs per component.}
\label{tab:latency}
\small
\begin{tabularx}{\columnwidth}{@{}X>{\centering\arraybackslash}p{0.19\columnwidth}>{\centering\arraybackslash}p{0.16\columnwidth}>{\centering\arraybackslash}p{0.16\columnwidth}@{}}
\toprule
\textbf{Component} & \textbf{Backend} & \textbf{p50 (s)} & \textbf{p95 (s)} \\
\midrule
Classifier (21-sentence reference essay) & HF (BF16) & 1.30 & 1.32 \\
Scorer (one essay) & CPU (32 threads) & 0.006 & 0.009 \\
\bottomrule
\end{tabularx}
\end{table}

\paragraph{GPU memory usage.}
The classifier (HF backend, BF16, Qwen2.5-7B + LoRA) uses a peak of
16.7 GB allocated GPU memory (18.5 GB reserved) for a single-essay
inference with 21 sentences. We do not report a memory or latency number for
the Qwen2.5-14B feedback backend because it was not benchmarked in this
release; consequently, the reported latency is not an end-to-end full-pipeline
latency.

\subsection{Summary of Experimental Findings}

The experiments yield four main findings. First, the logit-probe classifier
with Qwen2.5-7B + LoRA achieves 82.58\% accuracy and a macro-F1 of 0.727 on
the 12,000-sentence PERSUADE 2.0 test set, with claim--data confusion
accounting for 73\% of the errors. Second, the full 31-feature LightGBM
scorer achieves a mean QWK of 0.813 over 5-fold CV (fold range 0.754--0.872),
with gold discourse features contributing an increment of +0.055 QWK over the
lexical+syntactic configuration in the oracle component evaluation (paired
$t$-test, $p=0.010$). Third, the
scorer runs at sub-10-ms scorer-only latency on CPU, while the classifier
completes a 21-sentence essay in approximately 1.3 seconds on a single RTX
5000 Ada with a peak GPU memory of 16.7 GB. Fourth, the test suite passes
22/22 offline tests, supporting the reproducibility claims of the release.

\section{Open-Source Release Strategy}
\label{sec:release}

\begin{table}[h]
\centering
\caption{Asset distribution policy.}
\label{tab:assets}
\scriptsize
\begin{tabularx}{\columnwidth}{@{}p{0.25\columnwidth}p{0.34\columnwidth}X@{}}
\toprule
\textbf{Asset} & \textbf{Location} & \textbf{Reason} \\
\midrule
Source code, documentation, license & Git repository & Version control \\
Adapter config + tokenizer & Git repository (1.2 MB) & Small, user IP \\
LightGBM model + scaler & Git repository (33 MB) & Offline reproducibility \\
LoRA weights & GitHub Release (309 MB) & Avoids Git LFS bandwidth \\
Base models (Qwen2.5-7B, 14B) & HuggingFace Hub (Apache 2.0) & License restriction \\
PERSUADE 2.0 corpus & Upstream non-commercial license, \href{https://github.com/educational-technology-collective/persuade_2.0}{repository} & Non-commercial research use \\
\bottomrule
\end{tabularx}
\end{table}

\section{Testing and Quality Assurance}
\label{sec:testing}

The \texttt{tests/} suite validates eight aspects of the system, as listed
in Table~\ref{tab:tests}. The suite is intentionally offline: the vLLM API
contract test uses a fake engine rather than requiring a live model service.

\begin{table}[h]
\centering
\caption{Test coverage summary.}
\label{tab:tests}
\scriptsize
\begin{tabularx}{\columnwidth}{@{}p{0.30\columnwidth}X>{\centering\arraybackslash}p{0.14\columnwidth}@{}}
\toprule
\textbf{Test module} & \textbf{Covers} & \textbf{Pass rate} \\
\midrule
Batch outputs & ZIP structure, manifest, CSV &
3/3 \\
Classifier prompt & Prompt template, label parsing, shared context contract &
4/4 \\
Feedback bands & Band thresholds &
4/4 \\
Frozen assets & Adapter config, tokenizer, SHA-256 &
1/1 \\
Gradio contract & I/O schema (single + batch), branding &
4/4 \\
Model configuration & Config precedence, path resolution &
2/2 \\
Scorer contract & Feature order, scaler, flags, score decoder &
3/3 \\
vLLM API & API contract (fake engine) &
1/1 \\
\midrule
\textbf{Total} & & 22/22 \\
\bottomrule
\multicolumn{3}{@{}l}{\scriptsize All 22 tests pass without a GPU or network connection.}
\end{tabularx}
\end{table}

\section{Conclusion}
\label{sec:conclusion}

This paper presented ArguLens, an open-source modular system for
automated essay scoring and label-aware feedback. The central contribution
is a decoupled pipeline that separates discourse classification,
feature-based scoring, and feedback generation, allowing each component to
be developed, tested, and replaced independently.

Our experiments show that the logit-probe classifier achieves
82.58\% accuracy and a macro-F1 of 0.727 on the PERSUADE 2.0 test set.
The LightGBM scorer with all 31 features obtains a QWK of 0.813 over
5-fold CV. In the oracle component evaluation, adding gold discourse
features contributes an increment of +0.055 QWK over lexical and syntactic
features alone (paired $t$-test, $p=0.010$); this should not be interpreted as
the gain from predicted classifier features.
The feedback generator's evaluation protocol is defined in the repository;
a human-annotator study of feedback actionability is left to future work.
The reported component benchmarks are approximately 1.3 seconds for
classification and 6--9 ms for scoring; the feedback backend and therefore
full-pipeline latency were not benchmarked.

Extending the pipeline beyond English argumentative essays in a single
educational context --- using an XLM-RoBERTa backbone for discourse
classification and multilingual LLMs for feedback --- is a natural next
step, as is a larger-scale human evaluation of feedback actionability
across diverse learner populations. The code and release metadata are
provided under Apache 2.0; third-party models and data retain their upstream
licenses. The evaluation protocol and model artifacts are released to
facilitate community-driven extensions.

\section{Responsible Use and Limitations}
\label{sec:limitations}

The system is trained exclusively on US middle-school argumentative essays
from PERSUADE 2.0. Its performance should not be expected to generalise to
other genres, age groups, or languages without domain-specific fine-tuning.
The corpus reflects specific demographic and curricular contexts, and scores
may carry demographic biases. This release does not report gender,
race/ethnicity, or other subgroup metrics; a future study should use the
corpus metadata, report subgroup support and uncertainty, and evaluate both
score accuracy and feedback quality before any educational deployment.
Pending such analysis, the system is intended for research and assistive
classroom use only, not for high-stakes automated decisions.

The classifier is evaluated on completed essays and is not tested on partial
drafts. This matters for formative use: argument-mining models can be less
robust when context is incomplete, even when they perform well on completed
essays \citep{schaller-etal-2025-dont}. The scorer also uses a single holistic
1--6 target and does not model rater-level variation, a limitation when
interpreting borderline cases \citep{gaudeau-2025-beyond}.

A further methodological caveat is that the two released evaluation
protocols are not an end-to-end evaluation. The classifier is evaluated on an
essay-disjoint sentence-level split, whereas the reported scorer
cross-validation uses a feature table whose discourse fields are PERSUADE
gold annotations. Thus, the +0.055 QWK result is an oracle component
diagnostic, not a claim about predicted classifier features. A prompt-held-out
end-to-end rerun requires the licensed raw essay text and will be the next
evaluation step; comparisons against ordinal logistic regression, zero-shot
or generative prompting baselines, and fine-tuned encoder classifiers remain
future work.

The classifier benchmark used 16.7 GB allocated and 18.5 GB reserved VRAM on
one RTX 5000 Ada. The scorer itself runs on CPU, while the feedback backend
can call an OpenAI-compatible service; resource requirements for the local
14B feedback deployment were not measured. The LoRA weights must be
downloaded separately and verified against the SHA-256 checksum before first
use.

\section*{Acknowledgments}
This paper is a research output of the Digital Humanities and Glottometrics Lab at Fudan University.

\bibliography{references}

\appendix
\section{Reproducibility Checklist}
\label{app:checklist}
\begin{itemize}[leftmargin=*]
    \item \textbf{Code}: \url{https://github.com/wwrwbs/AI\_AWE} (Apache 2.0)
    \item \textbf{LoRA weights}: \url{https://github.com/wwrwbs/AI_AWE/releases/tag/v0.1.0}
    \item \textbf{Models}: Qwen2.5-7B-Instruct and Qwen2.5-14B-Instruct
    (HuggingFace Hub, Apache 2.0)
    \item \textbf{Data}: PERSUADE 2.0 (upstream non-commercial license; \href{https://github.com/educational-technology-collective/persuade_2.0}{data repository})
    \item \textbf{Environment}: frontend, vLLM, and training lockfiles
    (the three version-pinned requirement files)
    \item \textbf{Checksums}: \path{model_checksums.sha256}
    \item \textbf{Tests}: \texttt{pytest -q} (offline)
    \item \textbf{Launch}: \texttt{run\_all\_services.sh}
    (auto-detects virtual environments, GPUs, and ports)
    \item \textbf{LoRA training config}:
    \path{qwen_move_classifier/configs/qwen25_7b_move_lora_sft.yaml}
    \item \textbf{Split statistics}:
    \path{qwen_move_classifier/data/dataset_summary.json}
    (essay-level split with seed 42; zero overlap between splits)
\end{itemize}

\section{Full Feature List}
\label{app:features}

Table~\ref{tab:features} enumerates all 31 features used by the scorer,
grouped by family. Lexical-diversity variants are computed over all words
(\texttt{\_aw}), content words (\texttt{\_cw}), and function words
(\texttt{\_fw}).

\begin{table}[h]
\centering
\caption{The 31 scorer features.}
\label{tab:features}
\scriptsize
\begin{tabularx}{\columnwidth}{@{}p{0.36\columnwidth}X@{}}
\toprule
\textbf{Feature} & \textbf{Description} \\
\midrule
\multicolumn{2}{@{}l}{\textit{Lexical (16; TAALED and basic counts)}} \\
word\_count & Total word count \\
awl & Average word length \\
basic\_nfunction\_tokens & Function-word token count \\
lexical\_density\_tokens & Lexical density (content tokens / total tokens) \\
msttr50\_* & Moving-average segmental TTR, window 50 \\
mtld\_original\_* & Measure of Textual Lexical Diversity \\
mtld\_ma\_bi\_* & Bilinear moving-average MTLD variant \\
hdd42\_* & HD-D hypergeometric diversity, 42 draws \\
simple\_ttr\_fw & Simple type--token ratio (function words) \\
root\_ttr\_fw & Root TTR / Guiraud's index (function words) \\
\midrule
\multicolumn{2}{@{}l}{\textit{Syntactic (8; QuanSyn dependency metrics)}} \\
mdd & Mean dependency distance \\
ndd & Normalized dependency distance \\
mhd & Mean hierarchical distance \\
mtdl & Mean total dependency length \\
vk & Valency $k$ index \\
mtw & Mean tree width \\
hi & Head-initial ratio \\
mrd & Mean root distance \\
\midrule
\multicolumn{2}{@{}l}{\textit{Discourse (7; predicted in deployment, gold in oracle evaluation)}} \\
n\_sent & Sentence count \\
n\_claim / n\_data & Claim / evidence sentence counts \\
pct\_data & Share of \texttt{data} sentences (\%) \\
pct\_counterclaim & Share of \texttt{counterclaim} sentences (\%) \\
pct\_rebuttal & Share of \texttt{rebuttal} sentences (\%) \\
has\_counter\_rebut & Binary: essay contains both counterclaim and rebuttal \\
\bottomrule
\multicolumn{2}{@{}l}{\scriptsize * computed over all words (\_aw), content} \\
\multicolumn{2}{@{}l}{\scriptsize words (\_cw), and function words (\_fw).}
\end{tabularx}
\end{table}

\end{document}